\documentclass[conference]{IEEEtran}
\IEEEoverridecommandlockouts
\usepackage{cite}
\usepackage{amsmath,amssymb,amsfonts}
\usepackage[noend]{algorithmic}
\usepackage{graphicx}
\usepackage{textcomp}
\usepackage{xcolor}
\usepackage{booktabs}
\usepackage{multirow}
\usepackage{microtype}
\usepackage{hyperref}
\usepackage{cleveref}
\usepackage{algorithm}
\usepackage{subcaption}
\usepackage{flushend}

\def\BibTeX{{\rm B\kern-.05em{\sc i\kern-.025em b}\kern-.08em
    T\kern-.1667em\lower.7ex\hbox{E}\kern-.125emX}}
\begin{document}

\title{Markov Senior - Learning Markov Junior Grammars to Generate User-specified Content}

\author{\IEEEauthorblockN{Mehmet Kayra Oguz and Alexander Dockhorn}
\IEEEauthorblockA{\textit{Faculty of EECS} \\
\textit{Leibniz University Hannover}\\
Hannover, Germany \\
dockhorn@tnt.uni-hannover.de} 
}

\IEEEoverridecommandlockouts
\IEEEpubid{\makebox[\columnwidth]{ 979-8-3503-5067-8/24/\$31.00~\copyright2024 IEEE \hfill} 
\hspace{\columnsep}\makebox[\columnwidth]{ }}
\IEEEpubidadjcol

\maketitle

\begin{abstract}
Markov Junior is a probabilistic programming language used for procedural content generation across various domains. However, its reliance on manually crafted and tuned probabilistic rule sets, also called grammars, presents a significant bottleneck, diverging from approaches that allow rule learning from examples. In this paper, we propose a novel solution to this challenge by introducing a genetic programming-based optimization framework for learning hierarchical rule sets automatically. Our proposed method ``Markov Senior'' focuses on extracting positional and distance relations from single input samples to construct probabilistic rules to be used by Markov Junior. Using a Kullback-Leibler divergence-based fitness measure, we search for grammars to generate content that is coherent with the given sample. To enhance scalability, we introduce a divide-and-conquer strategy that enables the efficient generation of large-scale content. We validate our approach through experiments in generating image-based content and Super Mario levels, demonstrating its flexibility and effectiveness. In this way, ``Markov Senior'' allows for the wider application of Markov Junior for tasks in which an example may be available, but the design of a generative rule set is infeasible.
\end{abstract}


\section{Introduction}

Procedural Content Generation (PCG) stands as a vital tool in modern game design, enabling the efficient creation of diverse and coherent game content. Learning-based approaches within PCG are particularly significant, offering developers the ability to rapidly generate variations of existing samples. By leveraging machine learning techniques, these methods streamline game development processes and enhance the replayability of generated content-driven games.

A category of PCG algorithms, characterized by predefined rules and constraints, exemplifies both simplicity and effectiveness in content generation. The Markov algorithm~\cite{markov1954theory}, conceived by Andrey Andreyevich Markov, stands as a foundational algorithmic framework for probabilistic rule-based generation systems. Markov Junior, developed by Maxim Gumin \cite{MarkovJunior}, builds upon the Markov algorithm's core principles, expanding its flexibility and rule definitions to broaden its utility in PCG tasks. Nevertheless, the reliance of Markov Junior on user-defined rule sets limits its practicality.

To address this challenge, we introduce a machine learning-driven algorithm designed to acquire probabilistic rule sets for Markov Junior guided by a user-generated sample. The proposed model called ``Markov Senior'' aims to learn a grammar, which, when executed with Markov Junior, facilitates the generation of varied and cohesive content, aligning with the characteristics of the provided example.

The paper presents the following key contributions:
\begin{itemize}
    \item Introduction of a tile pattern relation concept, facilitating the description of positional and distance relationships among objects within the provided sample.
    \item Development of a genetic (evolutionary) algorithm tailored for generating Markov Junior rule sets.
    \item Implementation of efficiency optimizations in the training procedure, including a divide-and-conquer-based approach, to enable streamlined generation of large-scale content.
    \item Demonstration and validation of the proposed approach through the generation of both small-scale images and Super Mario levels.
\end{itemize}

In the following sections, we will first introduce the necessary preliminaries for our work (\Cref{sec:background}). In \Cref{sec:related-work}, we present a selection of related work that we drew inspiration from in the conception of our work. \Cref{sec:MarkovSenior} introduces, Markov Senior, the genetic programming model to learn Markov Junior grammars by example. It is followed by a demonstration of several use cases in \Cref{sec:evaluation}. Finally, we conclude our work in \Cref{sec:conclusion-future-work} and provide an overview of its limitations and opportunities for future work.

\section{Background}
\label{sec:background}

The following subsections provide a comprehensive overview of key concepts central to the paper's focus, including the Markov Algorithm, its extension Markov Junior, genetic algorithms, and evaluation metrics for generative models. While the former two will be reviewed to give an overview of the probabilistic programming language used in Markov Junior, the latter will later be utilized to learn Markov Junior grammars.

\begin{figure*}[t]
    \centering
    \begin{subfigure}[b]{0.28\textwidth}
        \centering
        \includegraphics[width=\textwidth]{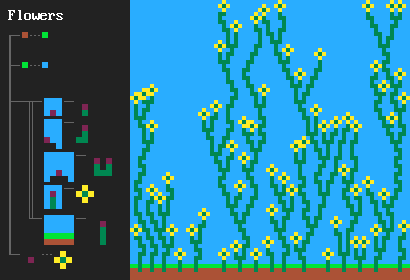}
        \caption{Flowers}
        \label{fig:MarkovJuniorFlowers}
    \end{subfigure}
    \hfill
    \begin{subfigure}[b]{0.28\textwidth}
        \centering
        \includegraphics[width=\textwidth]{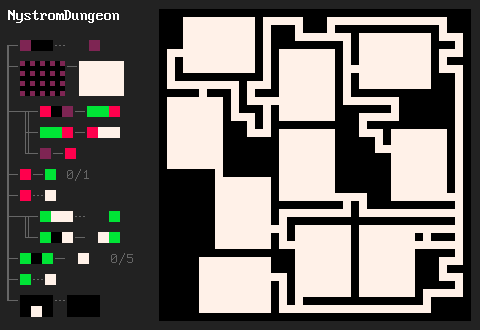}
        \caption{Dungeon}
        \label{fig:MarkovJuniorDungeon}
    \end{subfigure}
    \hfill
    \begin{subfigure}[b]{0.28\textwidth}
        \centering
        \includegraphics[width=\textwidth]{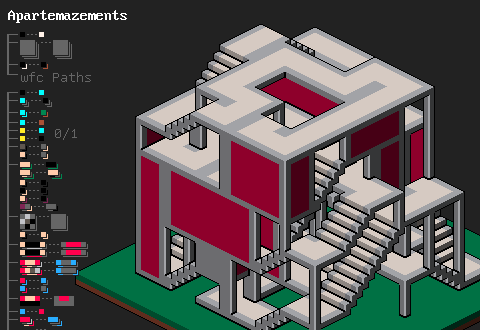}
        \caption{Architecture}
        \label{fig:MarkovJuniorArchitecture}
    \end{subfigure}
    \caption{Examples of Markov Junior grammars and their output taken from \cite{MarkovJunior}.}
    \label{fig:examples-markov-Junior}
\end{figure*}

\subsection{The Markov Algorithm and Markov Junior}
\label{sec:markov-algorithm}

The Markov Algorithm \cite{markov1954theory} is fundamentally a Turing-complete language model, modifying character strings through a hierarchical rule system where higher-level rules take precedence. Each rule comprises an antecedent and a consequent, representing substrings. During each iteration, the algorithm selects the first applicable rule, replacing the leftmost substring matching the antecedent with the consequent substring. Wildcards within substrings broaden their applicability, allowing for any symbol from the alphabet to be placed at the given position. The process of applying rules repeats until no more replacements on the string can take place or when a terminating rule is reached. The resulting string is the output of the algorithm.

The formal definition of the Markov Algorithm adopted in this paper is a 4-Tuple $MA = (\Sigma, R, G, E)$, consisting of an alphabet $\Sigma$, rules $R$, grammar $G$, and an environment $E$, where
\begin{itemize}
    \item $\Sigma$ is a finite and non-empty set of individual character strings.
    \item $R$ is a finite and non-empty set of character strings from $\Sigma$.
    \item $G$ is a finite, non-empty, and ordered set of rules from $R$.
    \item $E$ is a string, where $G$ is applied.
\end{itemize}

\subsection{Markov Junior}
\label{sec:MarkovJunior}

Inspired by the versatile capabilities of the Markov Algorithm, Maxim Gumin introduced Markov Junior~\cite{MarkovJunior}. Focusing on procedural content generation, this algorithm imposes additional constraints on a rule's definition, introduces an additional control structure called rule sets, and ensures their applicability to multi-dimensional inputs $E$. In handling multiple dimensions, where natural string insertion is not feasible, Markov Junior relies on constant-size replacements. Thus, both antecedents and consequences of rules in Markov Junior are defined to possess equal-size and -dimensioned substrings. Moreover, during rule application, multiple alternatives to applying a rule exist:
\begin{itemize}
    \item \textbf{One:} Replaces a single occurrence of the input string with the output string.
    \item \textbf{All:} Greedily choose and replace a subset of non-overlapping occurrences of the input string, with the output string.
    \item \textbf{Probabilistic:} Replaces all occurrences of the input string in a randomized order. Due to the overlap, the result will differ depending on the chosen order.
\end{itemize}

To bolster rule reusability, Markov Junior introduces rule set nodes. These can be stacked, enabling the construction of multi-level rule hierarchies. Two types of rule set nodes exist:
\begin{itemize}
    \item \textbf{Sequence Nodes:} encapsulate a series of rules or rule sets, which are applied sequentially for a fixed amount of iterations.
    \item \textbf{Markov:} Applies the Markov Algorithm to encapsulated rules and/or rule sets, i.e. execute the first applicable rule and repeat until no rule can be applied.
\end{itemize}

While the resulting process isn't Turing complete, it effectively describes a wide array of intriguing random processes, as demonstrated by examples from the developer's GitHub page~\cite{MarkovJunior} (see \Cref{fig:examples-markov-Junior}).

\subsection{Genetic Programming}

Genetic algorithms apply Darwinian principles of genetic evolution and represent a metaheuristic~\cite{holland1992adaptation}. They simulate crossover and mutation techniques on genome models within a predefined virtual environment governed by specific rules and laws. Starting with a randomized set of solution candidates, their objective is to generate offspring that meet predefined requirements as closely as possible. 

The branch of genetic programming algorithms~\cite{koza1994genetic} describes a technique to evolve programs often represented as trees. Nodes represent terminal symbols and operations to combine them. Throughout an evolutionary algorithm, these tree structures are adapted to solve a given task. Genetic algorithms provide a degree of randomized variation and development with crossover and mutation methods, however, these methods must be designed to be as much productive as possible, as they tend to cause information loss per generation when radical changes are forced on the genomes. Furthermore, small changes in the tree's structure can easily have a strong impact on the resulting output, putting a great emphasis on the design of suitable genetic operators.

\begin{figure*}[t]
    \centering
    \includegraphics[width=0.95\textwidth]{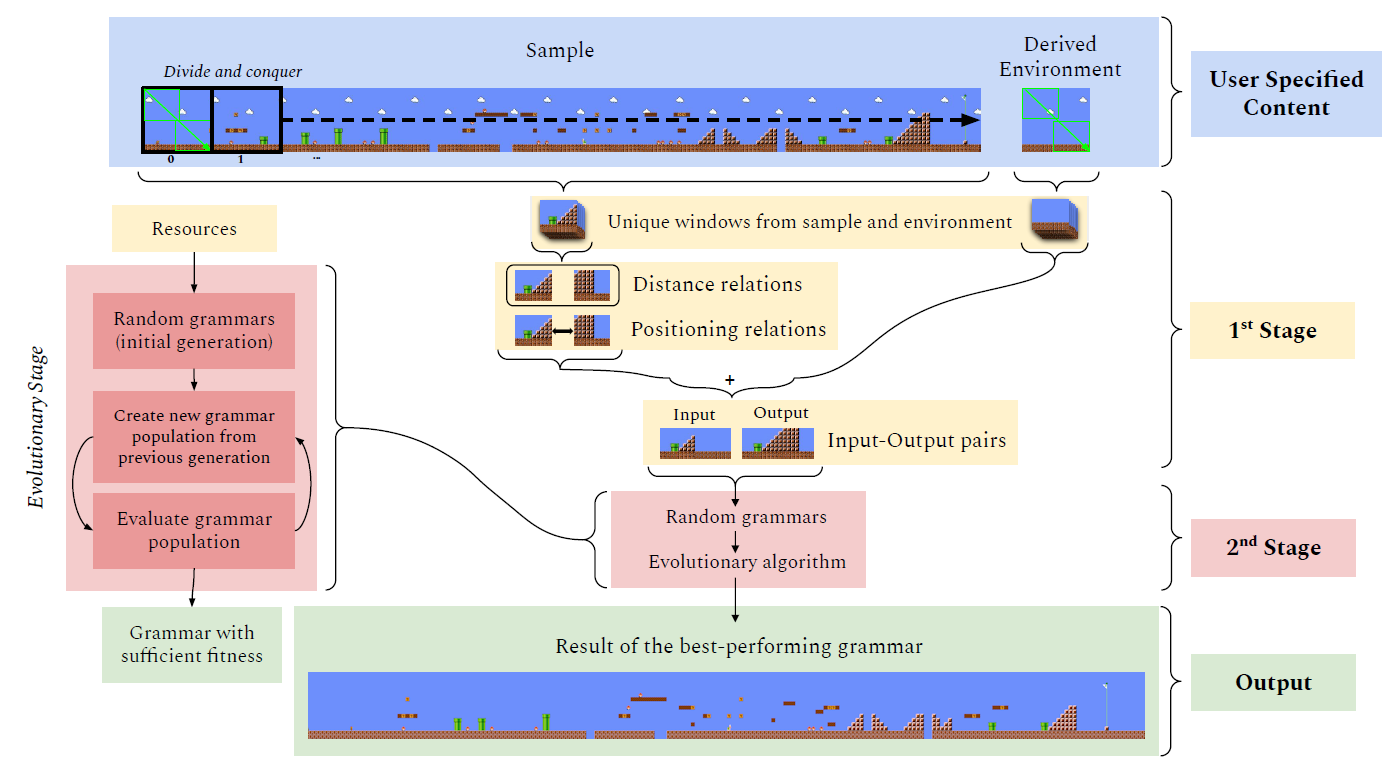}
    \caption{An overview of the whole evolutionary process.}
    \label{fig:ea-overview}
\end{figure*}

\subsection{Evaluation of Procedural Content Generation}

The evaluation of generated content in PCG is a hard-to-solve problem. Many works use statistical measures to compare the structure of a given example level with the output of a PCG algorithm. Lucas and Volz~\cite{kldivergence} have demonstrated the use of Kullback-Leibler (KL) divergence for the evaluation of Super Mario levels. KL-Divergence measures how much two probability distribution differs from each other. The KL-Divergence $D_{KL}$ of two probability distributions $P$ and $Q$ is defined by:
\begin{equation}
    D_{KL}(P||Q) = \sum_{x\in X} P(x) log (\frac{P(x)}{Q(x)})
\end{equation}
Applying it to compare pattern distributions allows us to measure the coherence of the given input and the generated output level. 

Minimal divergence is achieved by perfectly reproducing the input level, whereas random patterns often yield a high divergence. Generated outputs that keep substructures intact are able to achieve a low KL divergence. 

\section{Related Work on Sample-based PCG}
\label{sec:related-work}

Machine learning-based procedural content generation approaches, where a generator is trained by a user-specified example, is a widely researched field of PCG due to its usability. The popular method wave function collapse stems from the same author as Markov Junior~\cite{Gumin_Wave_Function_Collapse_2016}. It applies the idea of constraint satisfaction algorithms for PCG. Specifically, it reorders snippets from the sample with respect to a set of constraints originating from the same to generate new content. Thus, this technique generates new content that satisfies the obtained constraints. We will later derive this concept for the definition of probabilistic rules for Markov Junior.

A second work we drew inspiration from is the paper ``Linear levels through n-grams'' by Dahlskog et al.~\cite{dahlskog2014linear}. In their paper, they have shown that level generation for two-dimensional platformer games such as Super Mario can be enabled by extracting and replicating n-grams of vertical slices from a given example. In contrast to wave function collapse, this technique allows us to easily capture larger contexts due to the size of these n-grams. Working on substrings of arbitrary length, Markov Junior can potentially replicate this behavior. Nevertheless, the number of potential patterns and n-grams need to be constrained, which is why we will introduce a position and distance relation-based constraint for the selection of useful n-grams.

Next to such simple constraint- or pattern-based approaches, a large amount of work is focusing on the application of deep neural networks for content generation. While many of these approaches lack control over the resulting output, recent works on conditionalizing the output given an example~\cite{AwiSch2021} or based on text prompts~\cite{mariogpt} have achieved believable results. Nevertheless, the learned models are of black-box nature and hard to tune, configure or interpret.

With our line of work, we want to combine the sample efficiency of wave function collapse, which learns from a single example, with the expressivity of the probabilistic programming language Markov Junior. This combination aims to learn grammars that can later be tuned and adapted to the user's needs.

\section{Markov Senior - Learning Markov Junior Programs By Example}
\label{sec:MarkovSenior}

In this section, we present our genetic programming-based learning algorithm designed specifically for Markov Junior programs. The algorithm consists of two sequential stages: (1) preprocessing and (2)  evolutionary optimization. During the preprocessing stage, the provided example is analyzed to identify symbols, patterns, and their relationships. These insights are then utilized to initialize the evolutionary optimization of Markov Junior programs in the second stage, allowing to track their performance in generating coherent content aligned with the provided example. 

The following subsections will guide you through the multi-stage process and explain challenges and their solutions on the way. Finally, we will describe a divide-and-conquer-based optimization of the learning process to speed up the learning process of Markov Junior programs. To provide an initial overview and a reference while reading the subsections, the whole process has been summarized in \Cref{fig:ea-overview}.

\subsection{$1^{st}$ Stage - Preprocessing}

The preprocessing stage analyses the provided example and aims to prepare the evolutionary optimization stage.

\begin{figure}[t]
    \centering
    \includegraphics[height=4cm]{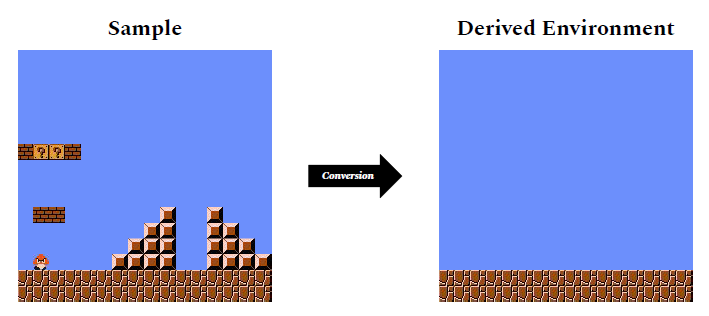}
    \caption{Initializing a Mario level based on the given sample.}
    \label{fig:initialization}
\end{figure}

First, we are going to infer the alphabet $\Sigma$ by keeping track of all symbols in the provided example. Given a picture, $\Sigma$ would encompass all the colors present in the image, given a tile map, we store the tile indices of included tiles. Furthermore, we need to derive the environment (output region) of the subsequent generative process. We can always choose to use an environment of the same size as the input example. Nevertheless, the rules that will later be learned can potentially be applied to environments of any size. Furthermore, we can choose to guide the following generation, by initializing the environment in a similar way as the provided example, e.g. reusing parts of a provided image or level. In our later experiments on Super Mario levels, we will be starting with a flat plane of ground tiles and filling the remainder with air tiles, therefore representing an ``empty'' level as represented in \Cref{fig:initialization}. Alternatively, we could start with a blank space. 

\begin{figure}[t]
    \centering
    \includegraphics[height=4cm]{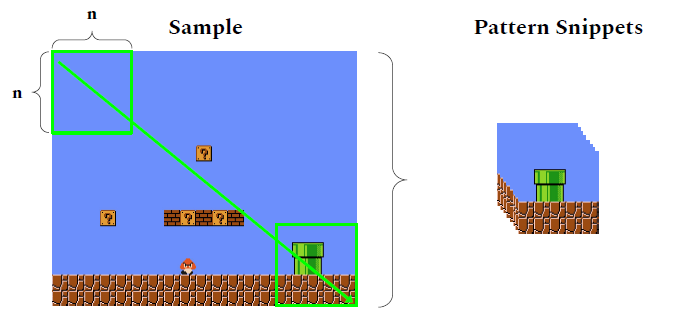}
    \caption{\textbf{$1^{\textit{st}}$ Stage - } Scanning a sample with an $n \times n$ window.}
    \label{fig:processing}
\end{figure}

During the generative process, we want to reproduce structures that occur in the given example level. Therefore, the rules' consequent should consist of patterns from the original level. Since those are produced by a series of rule applications, we also want the antecedent to consist of such patterns, such that they can be built step-by-step.
To extract potential antecedents and consequences for the rule-generation process, we will be scanning the level for patterns of fixed size. In case, an initialization level has been created, we need to scan it as well and add its patterns to the database. This ensures that new patterns that emerge from the initialization are covered and allow the generative process to start. \Cref{fig:processing} shows the pattern extraction process. For additional flexibility, tiles in any pattern, be replaced with a wild card. This allows for the generation of new patterns that have not been part of the sample level.

\subsubsection{Pattern Relations}

Next to the patterns' existence, we can gain much more information from the provided input level. More specifically, we aim to extract the patterns' distance and positioning relations. These relations provide information regarding relative local placements for each pattern pair combination. The distance relation investigates which two patterns should contextually be placed together, whereas the positioning relation investigates where exactly these patterns relative to each other may be placed.

\textbf{Pattern distance relations} 
%
provide information regarding contextual coherency. Meaning, they specify how far away two patterns could be from each other, to still represent an important combination. For each pattern of type $Q$ and the closest pattern of type $P$ with coordinates $q, p \in \mathbb{R}^2$, the distance relation $dr(Q, P)$ is defined by:
\begin{equation}
    dr(Q, P) = ||q - p||_2 = \sqrt{\sum_{i=1}^n(q_i-p_i)^2}.
\end{equation}
Next to considering the closest occurrence, we may want to filter based on the top $n$-closest occurrences, to reduce the impact of outliers.

\textbf{Pattern positioning relations }
%
provide information regarding their relative positional differences, hence indicating where exactly a pattern might be located relative to another. For each pattern $Q$ and the closest pattern of type $P$ with coordinates $q,p\in \mathbb{R}^2$, their positioning relation $pr(Q, P)$ is defined by:
\begin{equation}
    pr(Q, P) = q - p =(q_1 - p_1, q_2-p_2).
\end{equation}


\begin{figure}[t]
    \centering
    \begin{subfigure}[b]{0.45\textwidth}
        \includegraphics[width=1.0\textwidth]{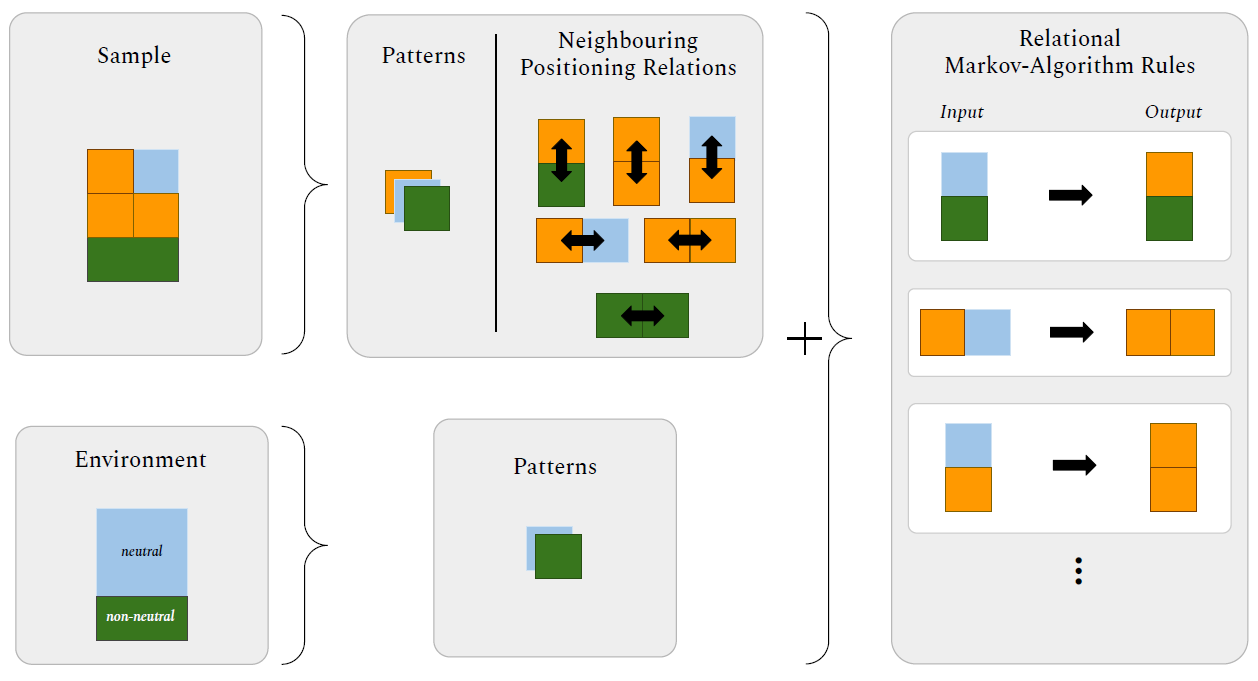}        
        \caption{Generation and application of relation-bounded rules.}
        \label{fig:relationalMarkovAlgorithmRules}
    \end{subfigure}
    \hfill
    \begin{subfigure}[b]{0.45\textwidth}
        \includegraphics[width=1.0\textwidth]{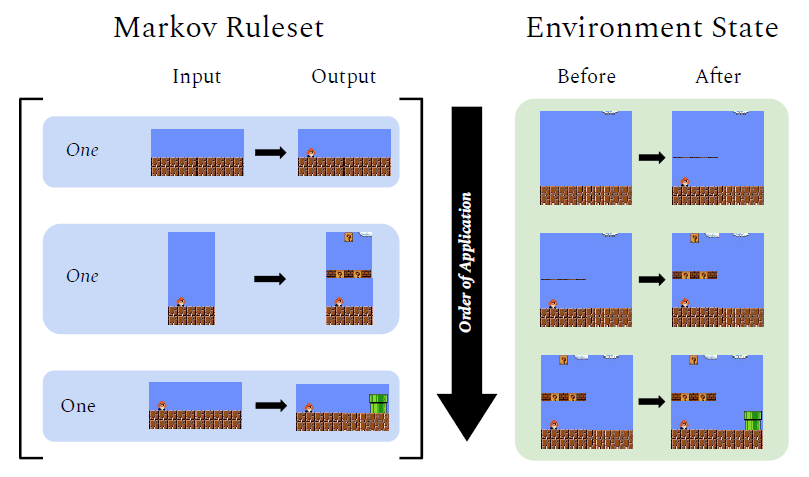}
        \caption{Step-by-step application process of a Markov rule set, which consists of one anchoring rule and two progressive rules.}
        \label{fig:markovruleset-application}
    \end{subfigure}
    \caption{\textbf{$1^{\textit{st}}$~Stage - } Pattern extraction, rule generation,\\ and application.}
    \label{fig:rule-application}
\end{figure}

\subsubsection{Creating Relation-bounded Rules}

Creating rules for Markov Junior is a complex task, due to the large number of patterns present in PCG environments. Given the exponential increase in the number of patterns based on the pattern's size, this problem quickly gets out of hand. For this reason, generating rules at random has a slim chance of ever finding a combination that is able to produce a meaningful level.

To increase the likelihood of generating rules that result in levels that are coherent with the provided sample, we make use of the defined distance and positioning relations. This allows us to pair collected pattern snippets based on constraints on their maximal distance and relative positioning. Hence a collection of relation-bounded rules is formed and modified over time in the algorithms $2^{nd}$ stage. 

A \textbf{relation-bounded rule} solely consists of an input-output pair that consists of patterns from the original level and follows additional constraints on the extracted positional and distance relations. First, we filter pattern combinations based on a maximal pattern distance relation, and second, we only generate input/output pairs for matching positioning relation offsets. For example, we only allow the current right neighbor of object $A$ to be replaced by anything that has also been observed to be the right neighbor of $A$ in the given example. Consequently, the applied grammar will potentially result in a series of modifications to the environment, which together compose an outcome that resembles the style of the sample from which the relations have been extracted. \Cref{fig:rule-application} demonstrates the pattern extraction, rule generation, and their application to a given environment.

In early experiments, we have observed two types of rules emerging from the generative process. We distinguish them by their function:
\begin{itemize}
    \item \textbf{Anchoring rules} attach to meaningful/non-neutral patterns in the initial environment (antecedent) and produce the starting points for the generation of more complex structures (rule consequent).
    \item \textbf{Progressive rules} build upon the results of anchoring rules or other progressive rules. With every iteration, they increase the affected area and help build more complex structures.
\end{itemize}

The neutrality of a pattern indicates its relative importance within the output. The determination of neutral and non-neutral patterns is context-specific and may be given by the user. In the exemplary scenario of generating Super Mario levels, air tiles can be considered neutral, whereas ground blocks are non-neutral. Enforcing this concept during grammar initialization (cf. \Cref{sec:generating-trees}), e.g. by including at least one non-neutral tile in a rule's antecedent and replacing only neutral tiles, may help to reduce the complexity of the grammar initialization and improve the efficiency of the evolutionary process.

\Cref{fig:markovruleset-application} shows the application of anchoring and progressive rules. Positioning relations that are composed of neutral and non-neutral tiles act as antecedents of anchoring rules. The composition of progressive rules is more versatile, but they require the preceding application of an anchoring rule.

\subsection{$2^{nd}$ Stage - Evolutionary Optimization}

In the second stage, we make use of an evolutionary optimization approach to learn Markov Junior grammars. During our optimization, a grammar tree represents the genotype and the generated output of said tree the respective phenotype. In the following subsections, we will first explain the representation and generation of grammar trees and how they can be genetically modified. Furthermore, we will describe their fitness evaluation in more detail.

\subsubsection{Representation and Generation of Markov Junior Grammars}
\label{sec:generating-trees}

\begin{figure}[t]
    \centering
    \includegraphics[width=0.49\textwidth]{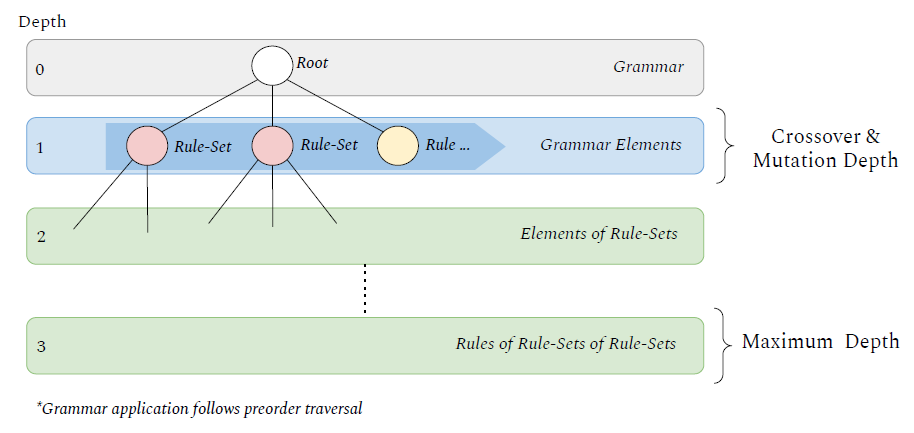}
    \caption{\textbf{$2^{\textit{nd}}$ Stage:} Genetic model of a Markov Junior grammar with its structural constraints.}
    \label{fig:grammar-tree}
\end{figure}

The genetic model used in Markov Senior consists of a tree-based representation of a Markov Junior grammar. Each inner node represents a rule set encapsulating its child nodes, whereas leave nodes represent singular rules. For rule nodes, we differentiate the rule types \textit{One} and \textit{All}. The \textit{Probabilistic}-node type is omitted, because of its increased computational overhead, which otherwise would slow down the optimization process. For rule set nodes we include both, \textit{Sequence} and \textit{Markov} nodes. To reduce the complexity of the optimization process, we limit the grammar tree to a maximum depth of 3 (not accounting for the root node). Furthermore, the root node is always a Sequence node, to ensure easier interpretability of the resulting process. The tree structure is depicted in \Cref{fig:grammar-tree}.

For the generation of grammar trees, we constrain our search space to the set of relation-bounded rules defined by a maximum distance for the pattern distance relation. For any triple of patterns $Q$, $P$, $P'$ for which it is known that $Q$ and $P$ have the same positional offset as $Q$ and $P'$ (as given by $pr(Q, P) = pr(Q, P')$), we generate a rule of type $Q,P_1 \rightarrow Q,P_2$. The antecedent includes pattern Q and P with the offset given in pr(Q,P), whereas the consequent includes pattern $Q$ and $P'$ with the same offset, effectively replacing the elements in $P$ with the ones in $P'$. When sampling rule nodes, we give nodes of type ``One'' a higher chance of being sampled, to motivate integral changes of the environment at the first iterations of optimization. 

Rule set nodes are randomly sampled by adding a random set of relation-bounded rules or previously created rule set nodes with a maximal depth of two to either a sequence rule set node or a Markov rule set node. Hence, allowing for a maximal depth of 3 in the generation process However, Markov nodes tend to become too complex if they include other Markov nodes in their respective trees. Therefore, Markov nodes can only reference rule nodes or sequence nodes, which include no Markov nodes, as their children. Furthermore, we limited rule set nodes to have a maximum of five child elements. These structural constraints ensure controlled genome growth and development in the evolutionary process but may be lifted if more computation resources are available.

\subsubsection{Genetic Modifications of Markov Junior Grammars}

Tree-based (hierarchical) genome models are potentially vulnerable to profound branch expansion during an evolutionary process~\cite{kaufmann2013adapting}. This implies drastic information loss per generation due to the hierarchical complexity accumulation with large numbers of deep branches. Therefore, crossover and mutation techniques must not cause radical changes in tree-based genome models. A conservative way to achieve this is to crossover only sub-trees that share similar depths and mutate only the sub-trees with a specific tree depth. Therefore, a series of rule sets and grammar-related constraints are forced on the grammar modification mechanisms.

\begin{figure}
    \centering
    \includegraphics[width=0.49\textwidth]{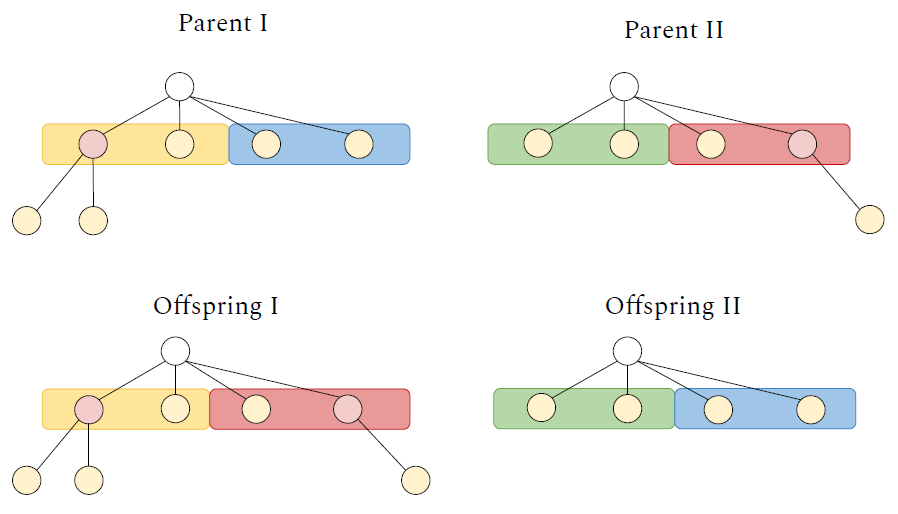}
    \caption{\textbf{$2^{\textit{nd}}$ Stage - } Crossover of two Markov Junior grammars and their offsprings.}
    \label{fig:crossover}
\end{figure}

We limit the application of crossover to the first level of the tree and swap the nodes of two Markov Junior grammar trees using a one-point crossover. Thus, for both trees, we select a cutoff point and build new candidate solutions by selecting the first half of the first parent's tree and the second half of the second parent's tree. \Cref{fig:crossover} visualizes the process. Specifically for sequence nodes, using a one-point crossover ensures that the two chosen subsequences remain intact. Furthermore, this crossover operator ensures fast exploration of the search space, by swapping large portions of the tree.
For the mutation of grammar trees, we implemented multiple operators of which a random one is applied to a random node of the grammar tree's first level. We either replace it with a random new node (rule or rule set node), delete it, or add a new sibling node. New nodes are randomly sampled from the set of relation-bounded rule or rule set nodes. 

\subsection{Measuring the Fitness}

For the evaluation of generated Markov Junior grammars, we rate the coherency of the generated outcome with the provided sample using pattern KL divergence~\cite{kldivergence}. To rate the fitness for any given pattern size of $n\times n$, we use:
\begin{equation}
    F(P,Q) = -(w\cdot D^{n}_{KL}(P\vert \vert Q) + (1-w) \cdot D^n_{KL} (Q \vert \vert P)
\end{equation}
for which $w$ represents a novelty factor that can ensure that the grammar does not overfit the given sample level. The resulting fitness measure is to be maximized. To measure differences on multiple levels of granularity, we recommend taking the average of multiple window sizes into account while evaluating the pattern distributions.

During the selection of individuals for reproduction, we use a biased version of roulette wheel selection. Therefore, individuals are randomly picked with a weighted probability correlating with their relative fitness score. Once the fitness values for all individuals have been determined we rescale the distribution to the range of $\lbrack 0, 1\rbrack$ and use the resulting values to calculate a relative fitness per individual. While this already prioritizes better candidates in the selection process, early experiments have shown that further biasing the result toward selecting the best grammar trees increases their impact on upcoming generations. Therefore, individuals with exceptionally high relative fitness of more than $0.8$ are always selected for reproduction and paired with another grammar chosen by roulette wheel selection. Once a pair of parents has been selected, the crossover is applied and both children are mutated. The process is repeated $n$ times, after which the best individuals from the parent and child generation are selected to form the next population. 

We terminate the evolutionary optimization once a candidate with a sufficient fitness score has been found or a maximum number of generations has been performed.

\subsection{Divide and Conquer}

The frequent iterations of the pattern-matching process used in Markov Junior can become a problem when large content is generated. While this computational overhead is feasible when generating a single outcome, learning to generate outcomes that are coherent with a given example will require us to evaluate many individuals throughout the optimization process. For this reason, we implemented a divide-and-conquer scheme to speed up the generation, application, and evaluation of subgrammars that can be combined to an overall grammar capable of generating content that is similar to a given level.

For this purpose, we divide the input level into multiple consecutive chunks. For each of these chunks, we create a separate process, optimizing a Markov Junior grammar that can produce outputs that are coherent with the given chunk. This reduces the run-time complexity of the pattern-matching and therefore the whole optimization. The outputs of the resulting Markov Junior grammars can later be stitched together to generate a level that is coherent with the whole level. The divide-and-conquer scheme is shown in the first row of \Cref{fig:ea-overview} and can be optionally applied to improve performance.

\begin{figure*}
    \centering
    \begin{minipage}{0.22\textwidth}
        \begin{subfigure}[c]{1.0\textwidth}
             \centering
             \includegraphics[height=1.5cm]{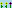}
             \caption{Flower Original Sample}
             \label{fig:original-flower}
        \end{subfigure}
        \vspace{0.5em}

        \begin{subfigure}[c]{1.0\textwidth}
             \centering
             \includegraphics[height=1.5cm]{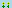}
             \caption{Flower Learned Grammar}
             \label{fig:generated-flower}
        \end{subfigure}
    \end{minipage}
    \hfill
    \begin{minipage}{0.76\textwidth}
        \begin{subfigure}[c]{0.49\textwidth}
             \centering
             \includegraphics[height=1.5cm]{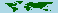}
             \caption{World Map Original Sample}
             \label{fig:world-map-original}
        \end{subfigure}
        \begin{subfigure}[c]{0.49\textwidth}
             \centering
             \includegraphics[height=1.5cm]{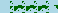}
             \caption{Learned World Map Grammar 1}
             \label{fig:world-map-1}
        \end{subfigure}
        \vspace{0.5em}

        \begin{subfigure}[c]{0.49\textwidth}
             \centering
             \includegraphics[height=1.5cm]{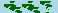}
             \caption{Learned World Map Grammar 2}
             \label{fig:world-map-2}
        \end{subfigure}
        \begin{subfigure}[c]{0.49\textwidth}
             \centering
             \includegraphics[height=1.5cm]{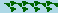}
             \caption{Learned World Map Grammar 3}
             \label{fig:world-map-3}
        \end{subfigure}
    \end{minipage}
    
    \caption{A collection of small-scale examples.}
    \label{fig:small-scale examples}
\end{figure*}

\begin{figure*}[t]
    \centering
    \begin{subfigure}[c]{1.0\textwidth}
         \centering
         \includegraphics[width=1.0\textwidth]{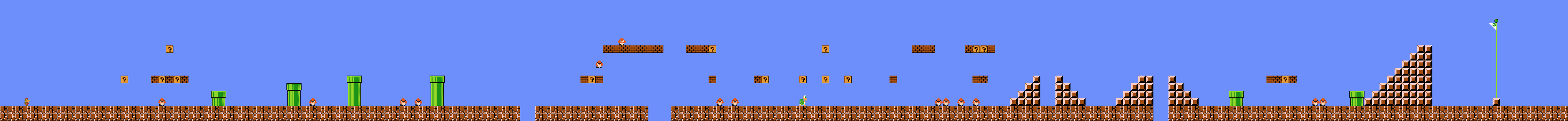}
         \caption{Original Super Mario Level 1-1}
    \end{subfigure}
    \vspace{0.5em}
    
    \begin{subfigure}[c]{1.0\textwidth}
         \centering
         \includegraphics[width=1.0\textwidth]{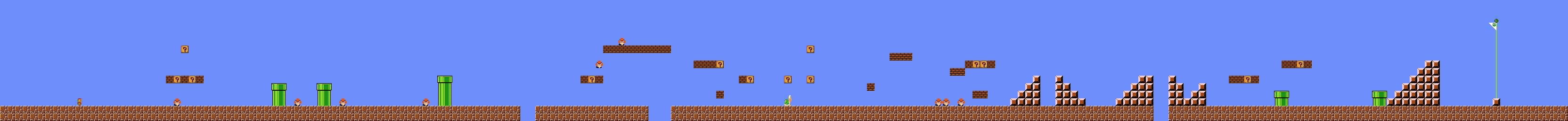}
         \caption{Learned Super Mario Level Grammar 1}
         \label{fig:good-results}
    \end{subfigure}
\vspace{0.5em}

     \begin{subfigure}[c]{1.0\textwidth}
         \centering
         \includegraphics[width=1.0\textwidth]{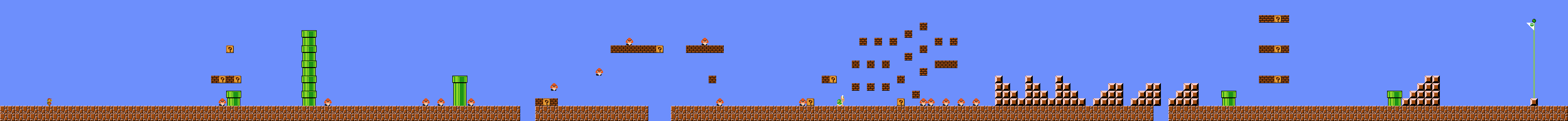}
         \caption{Learned Super Mario Level Grammar 3}
         \label{fig:bad-results}
    \end{subfigure}
    \vspace{0.5em}
    
    \begin{subfigure}[c]{1.0\textwidth}
         \centering
         \includegraphics[width=1.0\textwidth]{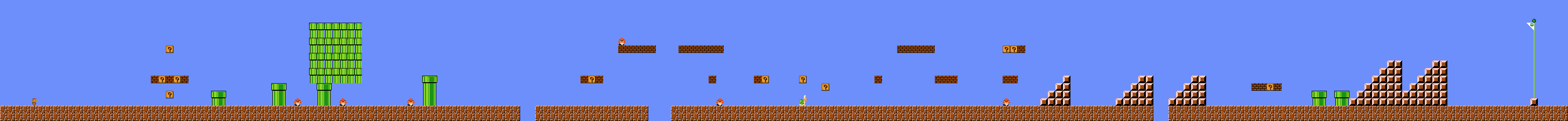}
         \caption{Learned Super Mario Level Grammar 2}
         \label{fig:clear-errors}
    \end{subfigure}

    \caption{Learning to generate larger content using the proposed divide-and-conquer scheme and a pattern size of $6\times6$.}
    \label{fig:divide-and-conquer}
\end{figure*}

\section{Demonstration of Markov Senior}
\label{sec:evaluation}

To demonstrate the learning capabilities of Markov Senior, we learn Markov Junior grammars for several examples. This way, we can investigate the visual characteristics of the generated content and test what the optimization algorithm is capable of when trained with contents that have different complexity. At this stage, we would like highlight that these test cases all focus on replicating the original content as best as possible to demonstrate the algorithm's learning capabilities. In future work, we aim to make this process more controllable by allowing users to choose between accuracy and novelty during the generation process. Since this level of control is not yet achievable with tested metrics, we will concentrate on this aspect in future studies.

\Cref{fig:small-scale examples} shows two small-scale examples for learning to generate content that is coherent with a given sample.  As the sample inputs are small enough for an acceptable runtime complexity, we learn a Markov Junior grammar to generate them as a whole. While the generated outputs follow the general style of the input image, we see small variations in the generated output images. As seen in the \Cref{fig:generated-flower} the generated flowers mostly follow their given counterparts, with the exception of one flower having two leaves which is not observed in the sample. Moreover, it can be observed in \Cref{fig:world-map-1,fig:world-map-2,fig:world-map-3} that generated patterns follow island-like appearances which resemble the continents in the original sample.

Motivated by the algorithm's small-scale capabilities, we tested it for the generation of Super Mario levels, which are of a much larger scale than the previous examples. Here, we identified the previously discussed bottlenecks in pattern matching, which led to the design of the divide-and-conquer approach used for the following examples. Due to splitting the input into multiple chunks and processing them separately, the final output will follow a similar style, but the grammar of each chunk may deviate in quality. \Cref{fig:divide-and-conquer} shows the original Super Mario level 1-1 and grammars learned by Markov Senior with varying accuracy in reproducing the result. Overall we can see that most substructures can be accurately reproduced (\Cref{fig:good-results}) but the optimization approach does not guarantee to find a good grammar in all runs (\Cref{fig:bad-results}). Some runs may even produce visible artifacts, such as the stack of pipes in \Cref{fig:clear-errors}. Nevertheless, those have often been resolved by restarting the training process or replacing the given chunk's grammar with the grammar from another run.

Overall, Markov Senior is capable of learning Markov Junior grammars for generating content of different complexity. Nevertheless, further studies will be necessary to study its capabilities and limitations as well as to allow users to control its output.

\section{Conclusion and Future Work}
\label{sec:conclusion-future-work}

It has been shown that Markov Senior, an evolutionary optimization algorithm for generating Markov Junior grammars, is capable of producing content that is coherent with a given sample. The extraction of relation-bounded rules has been introduced to reduce the search space during the grammar construction and optimization process considerably. Furthermore, a divide-and-conquer scheme has been designed to increase the efficiency of the optimization process. The code of our proposed method and presented examples can be found at: \textit{https://github.com/ADockhorn/MarkovSenior}.

While the results of our tested use cases look promising, we are aware of the study's limitations. For now, we do not have a comprehensive understanding of the impact of each of the algorithm's components. Our work has been driven by the results of initial experiments that largely failed to generate anything meaningful. Careful analysis of the problems that occurred during these failed runs has helped us to finalize this first version of Markov Senior. Further optimization of its fitness function, genetic operators, or even the representation and generation of grammars may show much better and more consistent results. However, to the best of our knowledge, this is the first work that proposes to learn Markov Junior grammars based on a given example. Therefore, we believe that its current state will already prove useful to the PCG community and spawn new ideas and methods for improving our design.

In the future, we aim to improve further the algorithm's learning speed and the user's control over the generated output. For this purpose, we aim for a detailed analysis of the method's parameter space and optimizations in the way grammars are genetically modified. Further optimization in the representation, mutation or crossover operators may be required to increase the performance of the grammar optimization~\cite{Doc2022a}. 

To boost the user's control over the outcome, we currently experiment with parameterizing the fitness function. The version hosted on Github introduces additional parameters to steer the tradeoff between novelty and coherence. While changes in the output become visible for extreme parameter settings, they do not yet allow fine-grained control over the learned grammars and their output. Nevertheless, this line of work will be continued and project updates will be shared when ready.

Finally, to follow recent trends in procedural content generation, one interesting addition would be the use of large language models for generating or processing grammars. Due to their power they see more frequent use in coding tasks as well as procedural content generation settings~\cite{AwiDoc2023}.

\bibliographystyle{plain}
\bibliography{references}

\begin{thebibliography}{10}

\bibitem{AwiDoc2023}
Maren Awiszus, Alexander Dockhorn, Amy~K. Hoover, Antonios Liapis, Simon~M. Lucas, Mirjam~Palosaari Eladhari, Jacob Schrum, and Vanessa Volz.
\newblock Language models for procedural content generation.
\newblock {\em Human-Game AI Interaction (Dagstuhl Seminar 22251)}, 12(6):34--37, January 2023.

\bibitem{AwiSch2021}
Maren Awiszus, Frederik Schubert, and Bodo Rosenhahn.
\newblock World-gan: a generative model for minecraft worlds.
\newblock In {\em IEEE Conference on Games}, August 2021.

\bibitem{dahlskog2014linear}
Steve Dahlskog, Julian Togelius, and Mark~J Nelson.
\newblock Linear levels through n-grams.
\newblock In {\em Proceedings of the 18th International Academic MindTrek Conference: Media Business, Management, Content \& Services}, pages 200--206, 2014.

\bibitem{Doc2022a}
Alexander Dockhorn.
\newblock Choosing representation, mutation, and crossover in genetic algorithms.
\newblock {\em IEEE Computational Intelligence Magazine}, 17(4):52--53, November 2022.
\newblock This is an immersive article. Therefore, extended interactive resources are provided at the publisher's webpage. A preprint can be found at: https://aiexplained.github.io/.

\bibitem{Gumin_Wave_Function_Collapse_2016}
Maxim Gumin.
\newblock {Wave Function Collapse Algorithm}, September 2016.
\newblock \url{https://github.com/mxgmn/WaveFunctionCollapse}.

\bibitem{MarkovJunior}
Maxim Gumin.
\newblock {MarkovJunior, a probabilistic programming language based on pattern matching and constraint propagation}, June 2022.
\newblock \url{https://github.com/mxgmn/MarkovJunior}.

\bibitem{holland1992adaptation}
John~H Holland.
\newblock {\em Adaptation in natural and artificial systems: an introductory analysis with applications to biology, control, and artificial intelligence}.
\newblock MIT press, 1992.

\bibitem{kaufmann2013adapting}
Paul Kaufmann.
\newblock {\em Adapting Hardware Systems by Means of Multi-Objective Evolution}.
\newblock Logos Verlag Berlin GmbH, 2013.

\bibitem{koza1994genetic}
John~R Koza et~al.
\newblock {\em Genetic programming II}, volume~17.
\newblock MIT press Cambridge, 1994.

\bibitem{kldivergence}
Simon~M Lucas and Vanessa Volz.
\newblock Tile pattern kl-divergence for analysing and evolving game levels.
\newblock In {\em Proceedings of the Genetic and Evolutionary Computation Conference}, pages 170--178, 2019.

\bibitem{markov1954theory}
Andrei~Andreevich Markov.
\newblock The theory of algorithms.
\newblock {\em Trudy Matematicheskogo Instituta Imeni VA Steklova}, 42:3--375, 1954.

\bibitem{mariogpt}
Shyam Sudhakaran, Miguel González-Duque, Claire Glanois, Matthias Freiberger, Elias Najarro, and Sebastian Risi.
\newblock Mariogpt: Open-ended text2level generation through large language models, 2023.

\end{thebibliography}

\end{document}